\documentclass[letterpaper]{article} 
\usepackage{aaai25}  
\usepackage{times}  
\usepackage{helvet}  
\usepackage{courier}  
\usepackage[hyphens]{url}  
\usepackage{graphicx} 
\usepackage{natbib}  
\usepackage{caption} 
\usepackage{algorithm}
\usepackage{algorithmic}
\usepackage{amsmath}
\usepackage{amssymb}
\usepackage{booktabs}
\usepackage{multirow}
\usepackage{multicol}
\usepackage{colortbl}
\usepackage{color}
\usepackage{makecell}
\usepackage{subfig}
\usepackage{xcolor}

\usepackage{newfloat}
\usepackage{listings}
\DeclareCaptionStyle{ruled}{labelfont=normalfont,labelsep=colon,strut=off} 
\floatstyle{ruled}
\newfloat{listing}{tb}{lst}{}
\floatname{listing}{Listing}
\title{Micro-macro Wavelet-based Gaussian Splatting for 3D Reconstruction from Unconstrained Images}
\author{
    Yihui Li\textsuperscript{\rm 1,2},
    Chengxin Lv\textsuperscript{\rm 1,2},
    Hongyu Yang\textsuperscript{\rm 3,4*},
    Di Huang\textsuperscript{\rm 1,2}
}
\affiliations{
    \textsuperscript{\rm 1} State Key Laboratory of Complex and Critical Software Environment, Beijing, China \\
    \textsuperscript{\rm 2} School of Computer Science and Engineering, Beihang University, China\\

    \textsuperscript{\rm 3} School of Artificial Intelligence, Beihang University, China\\
    \textsuperscript{\rm 4} Shanghai Artificial Intelligence Laboratory, Shanghai, China\\
    \{kidleyh, 20231195, hongyuyang, dhuang\}@buaa.edu.cn
}

\usepackage{bibentry}

\begin{document}

\maketitle

\begin{abstract}
3D reconstruction from unconstrained image collections presents substantial challenges due to varying appearances and transient occlusions. In this paper, we introduce Micro-macro Wavelet-based Gaussian Splatting (MW-GS), a novel approach designed to enhance 3D reconstruction by disentangling scene representations into global, refined, and intrinsic components. The proposed method features two key innovations: \textbf{Micro-macro Projection}, which allows Gaussian points to capture details from feature maps across multiple scales with enhanced diversity; and \textbf{Wavelet-based Sampling}, which leverages frequency domain information to refine feature representations and significantly improve the modeling of scene appearances. Additionally, we incorporate a Hierarchical Residual Fusion Network to seamlessly integrate these features. Extensive experiments demonstrate that MW-GS delivers state-of-the-art rendering performance, surpassing existing methods.
\end{abstract}

%

\section{Introduction}
3D reconstruction from images is a longstanding and critical task in computer vision, with applications ranging from immersive virtual reality to 3D content creation. Recent advancements in this field have been driven by both implicit representations, such as Neural Radiance Fields (NeRF) \cite{mildenhall2021nerf} and their subsequent developments \cite{barron2023zip, tancik2022block}, and explicit representations, such as 3D Gaussian Splatting (3DGS) \cite{kerbl20233d} and related techniques \cite{yu2024mip, lu2024scaffold}. These methods have demonstrated significant progress, particularly in reconstructing static scenes under controlled conditions, where stable lighting and minimal occlusions prevail. However, real-world applications often involve images captured in unconstrained and dynamic environments, where traditional methods struggle to maintain consistent reconstruction quality, leading to issues such as blurriness, artifacts, and overall performance degradation \cite{yang2023cross}.

To address the challenges of dynamic appearance variations in real-world scenes, NeRF-W \cite{martin2021nerf-w} introduced per-image appearance embeddings, later refined by \cite{chen2022hallucinated, yang2023cross} to better handle these variations across different views. Despite these advancements, global embeddings often fall short in representing fine-grained details, as they fail to fully account for the significant appearance variations across different scene locations influenced by object properties and environmental factors. Furthermore, the high computational costs associated with implicit representations hinder their applicability in real-time rendering scenarios.

Some Gaussian-related methods \cite{dahmani2024swag, kulhanek2024wildgaussians} have attempted to improve dynamic appearance modeling by combining intrinsic Gaussian features with global appearance embeddings. However, these approaches still face limitations similar to those of NeRF-based methods, primarily due to the less expressive nature of global embeddings. The state-of-the-art GS-W method \cite{zhang2024gaussian-w} has made notable progress by allowing Gaussian points to adaptively sample detailed dynamic appearance information from 2D feature maps. This technique provides more flexibility and captures richer details, yet challenges like blurriness persist, particularly when images are closely inspected.


In this paper, we introduce Micro-macro Wavelet-based Gaussian Splatting (MW-GS), a novel approach designed to address the limitations of existing 3D reconstruction techniques. Our method decomposes Gaussian features into three distinct components: global appearance, refined appearance, and intrinsic features, which together provide a comprehensive representation of dynamically changing scenes. Global features capture the overall scene characteristics, such as color tone and lighting conditions, while refined features model detailed textures and region-specific effects like highlights and shadows. Intrinsic features represent consistent properties of the scene, ensuring robustness across varying conditions.

The key innovation of our work lies in \textbf{Micro-macro Projection}, which significantly enhances refined appearance modeling-a critical aspect where previous methods often fall short. 
Specifically, our method employs adaptive sampling within narrow and broad conical frustums on the 2D feature map, optimizing the 3D Gaussian points to capture both fine-grained textures and relatively long-range features that influence overall scene fidelity, such as highlights. 
Inspired by traditional MipMap operations, we further introduce a simple yet effective jitter to the projected position for each Gaussian point at the \textbf{micro scale}, rather than relying on a fixed position as in previous methods. This jitter introduces variability into the sampling process, enabling capturing a richer set of features. 
Additionally, we incorporate \textbf{Wavelet-based Sampling}, leveraging frequency domain information to further enhance refined appearance modeling and reconstruction accuracy. This multi-scale approach ensures that each Gaussian point accurately captures fine-grained features while maintaining feature diversity. Finally, we propose a Hierarchical Residual Fusion Network (HRFN) to effectively integrate features from different levels, ensuring a cohesive and precise 3D reconstruction.

In summary, the key contributions of this paper are:

\begin{itemize}
\item Micro-macro Wavelet-based Gaussian Splatting (MW-GS): A novel approach for 3D reconstruction that effectively decomposes and models global, refined, and intrinsic scene features.
\item Micro-macro projection: An innovative sampling strategy that introduces jitter and adaptive conical frustums to capture a rich set of features, significantly enhancing refined appearance modeling.
\item Wavelet-based Sampling: This component enhances multi-resolution sampling, leveraging frequency domain information to refine feature representation and improve reconstruction accuracy.
\end{itemize}

Our experiments across multiple scenes demonstrate that MW-GS achieves superior reconstruction results and rendering quality, surpassing existing methods.

\section{Related Work}
\subsection{Scene Representations}
Existing traditional methods include meshes \cite{groueix2018papier, kanazawa2018learning, kato2018neural, wen2019pixel2mesh}, point clouds \cite{qi2017pointnet, qi2017pointnet++, shi2020pv}, and voxels \cite{schwarz2022voxgraf, wu20153d}. Recently, Neural Radiance Fields (NeRF) \cite{mildenhall2021nerf} have revolutionized the synthesis of novel, photo-realistic views from images. Extensions to NeRF enhance visual quality \cite{barron2022mip, hu2023tri}, rendering speed \cite{chen2023mobilenerf, reiser2023merf}, and convergence \cite{muller2022instant, chen2022tensorf}, though limitations persist in speed and detail.
More recently, 3D Gaussian Splatting (3DGS) \cite{kerbl20233d}, an explicit representation method, offers real-time rendering with high-resolution quality. Recent advances in 3DGS include improvements in efficiency \cite{lee2024compact}, surface reconstruction \cite{huang20242d}, and incorporating semantic attributes for multimodal applications \cite{xie2024physgaussian, shi2024language, qin2024langsplat}. 3DGS has also been extended to various tasks, including autonomous driving \cite{zhou2024drivinggaussian, yan2024street, zhou2024hugs}, 3D generation \cite{chen2024text, chung2023luciddreamer}, and controllable 3D scene editing \cite{chen2024gaussianeditor, wang2024gaussianeditor, zhou2024feature}.

\subsection{Novel View Synthesis from Unconstrained Photo Collections}
Traditional novel view synthesis methods assume static geometry, materials, and lighting conditions. However, unconstrained photo collections, such as those from the internet \cite{snavely2006photo}, often feature variable lighting and dynamic occlusions, which challenge these assumptions. NeRF-W \cite{martin2021nerf-w} pioneered addressing these challenges by incorporating learnable appearance and transient embeddings for each image. Subsequent NeRF-based methods, such as Ha-NeRF \cite{chen2022hallucinated} and CR-NeRF \cite{yang2023cross}, extended this approach with CNN-based appearance encoders to handle dynamic variations more effectively.

Despite these advancements, methods based on implicit representations suffer from slow rendering speeds, prompting research into the use of 3D Gaussian Splatting (3DGS) as a more efficient alternative to NeRF. Approaches such as SWAG \cite{dahmani2024swag} and WildGaussians \cite{kulhanek2024wildgaussians} incorporate learnable appearance embeddings to modulate the color of 3D Gaussians using a multi-layer perceptron (MLP). WE-GS \cite{wang2024we} enhances CNN-based appearance representations through a spatial attention module, while GS-W \cite{zhang2024gaussian-w} and Wild-GS \cite{xu2024wild} leverage CNNs to generate feature maps for dynamic appearance modulation. GS-W projects each Gaussian onto feature maps and performs adaptive sampling, whereas Wild-GS uses depth information to construct triplane maps for local appearance embedding.

However, projecting 2D appearance reference images to 3D space can lead to sparse features and gaps. In this work, we address these limitations by introducing Micro-macro Wavelet-based Sampling, which enhances sampling accuracy and diversity. Our approach leverages frequency domain information to refine appearance features, representing the first integration of frequency domain data into 3DGS appearance representation. This innovation effectively improves both the appearance representation and reconstruction results for unstructured photo collections.

\begin{figure*}[t]
\centering
\includegraphics[width=0.98\textwidth]{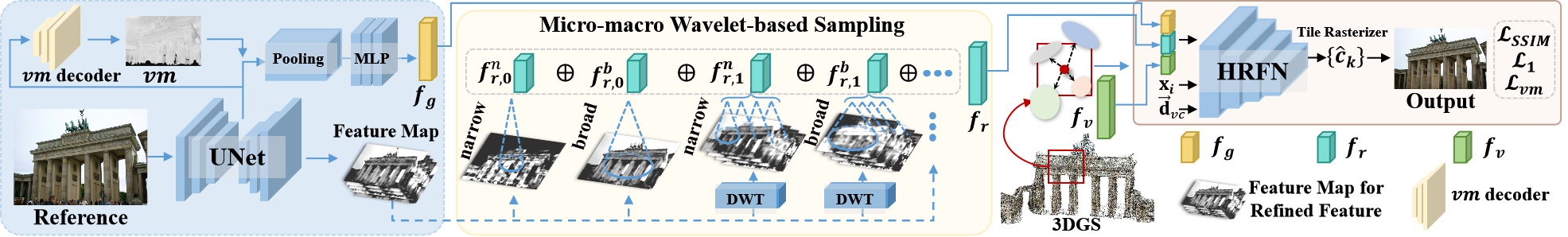}
\caption{Overview of Micro-macro Wavelet-based Gaussian Splatting (MW-GS). Beginning with a reference image, CNN-based sub-networks extract global appearance features, a visibility map, and a set of feature maps which then undergo wavelet transform to capture multi-resolution information for refined appearance modeling. 3D Gaussian points are projected onto these feature maps using Micro-macro Projection, which employs adaptive sampling within narrow and broad conical frustums to capture richer information. The global, refined, and intrinsic features are subsequently fused through a Hierarchical Residual Fusion Network (HRFN) to generate the final color, which is rendered using a tile rasterizer for high-quality 3D reconstruction.}
\label{fig:pipeline}
\end{figure*}

\section{Preliminaries}
\subsection{3D Gaussian Splatting}

3D Gaussian Splatting (3DGS) \cite{kerbl20233d} represents a scene using a collection of anisotropic 3D Gaussians. These Gaussians are differentiable and can be efficiently projected onto 2D splats, enabling rapid $\alpha$-blending through tile-based rasterization.
Each 3DGS is characterized by a complete 3D covariance matrix $\mathbf{\Sigma}\in \mathbb{R}^{3\times 3}$, which is defined in world space and centered at a point (mean) $\mu\in \mathbb{R}^3$:
\begin{equation}
    G(x) = exp(-\frac{1}{2}(x-\mu)^\top\mathbf{\Sigma}^{-1}(x-\mu)),
\end{equation}
where $x$ is an arbitrary position within the 3D scene. To maintain its positive semi-definiteness during optimization, \(\mathbf{\Sigma}\) is formulated using a diagonal scaling matrix \(\mathbf{S}\) and an orthogonal rotation matrix \(\mathbf{R}\):
\begin{equation}
    \mathbf{\Sigma} = \mathbf{R}\mathbf{S}\mathbf{S^\top}\mathbf{R^\top}.
\end{equation}
In terms of implementation, the covariance matrix $\mathbf{\Sigma}$ is parameterized using a unit quaternion $q$ and a 3D scaling vector $s$. Additionally, each Gaussian is associated with color $\hat{c}$ and an opacity factor $\alpha$, which is multiplied by $G(x)$ during the blending process. During rendering, these Gaussians are directly projected to the screen for high-speed rendering, called Splatting \cite{zwicker2001ewa}. After sorting the Gaussians by depth, alpha compositing is used to compute the final color $\widehat{C}$ for pixel $\mathbf{p}$. This process is expressed as:
\begin{equation}
\label{eq:alpha-blending}
    \widehat{C}(\mathbf{p}) = \sum_{i\in N} \hat{c}_i\alpha'_i\prod^{i-1}_{j=1}(1-\alpha'_j),
\end{equation}
where \(\alpha'_i\) is the product of \(\alpha_i\) and the splatted 2D Gaussian.

\subsection{Discrete Wavelet Transform}

Wavelet \cite{daubechies1992ten, strang1996wavelets}, has long been a widely used mathematical tool in image analysis \cite{chen2021local, duan2017sar, li2022wavelet}. The coefficients of the Discrete Wavelet Transform (DWT) describe signals across different frequency bands at various resolution levels, effectively capturing both local and global information.
The 2D DWT decomposes an image into four different components in the frequency domain with low-pass (emphasizing relatively smooth parts) and high-pass (capturing high-frequency signals, including local texture information) filters, denoted by $\mathbf{L}$ and $\mathbf{H}$, respectively.
The combination of the two filters yields four distinct kernels, \textit{i.e.}, $\mathbf{LL}$, $\mathbf{LH}$, $\mathbf{HL}$, and $\mathbf{HH}$, corresponding to different frequency and spatial information. Given a feature map $\mathbf{F}\in \mathbb{R}^{H\times W}$, where $H$ and $W$ indicate its height and width, respectively, we apply DWT to $\mathbf{F}$. The four sub-band features can be obtained by a one-level decomposition, formulated as:
\begin{equation}
\label{eq:wavelet}
\begin{aligned}
    \mathbf{F}_w^{LL} &=&\textbf{LF}\textbf{L}^\top,
    \mathbf{F}_w^{LH} &=&\textbf{HF}\textbf{L}^\top,
    \\
    \mathbf{F}_w^{HL} &=&\textbf{LF}\textbf{H}^\top,
    \mathbf{F}_w^{HH} &=&\textbf{HF}\textbf{H}^\top.
\end{aligned}
\end{equation}
For a feature map with multi channels, the wavelet transform is applied individually to each channel. The resulting sub-bands from the same filter are then concatenated, producing four corresponding frequency sub-bands.

\section{Method}
To achieve high-fidelity 3D reconstruction form unconstrained image collections, we introduce a novel approach named Micro-macro Wavelet-based Gaussian Splatting that decouples appearance attributes into global, refined, and intrinsic features, achieving a more structured and explicit scene representation. Our method consists of the following key components:
\textit{i)} Global appearance features are extracted directly from a 2D reference image using CNN-based networks.
\textit{ii)} We propose a Micro-Macro Wavelet-based sampling technique to enhance the accuracy and diversity of refined appearance features by capturing multi-resolution information.
\textit{iii)} A Hierarchical Residual Fusion Network (HRFN) effectively integrates learnable intrinsic feature embeddings with global and refined features.
\textit{iv)} Visibility maps are employed to mitigate the influence of transient objects during the optimization process.
This approach enables high-quality 3D reconstruction with greater attention to refined detail and overall appearance fidelity.

\subsection{Structured and Explicit Appearance Decoupling}

In unconstrained photo collections, global appearance variations arise from factors like lighting conditions during capture and post-processing techniques such as gamma correction, exposure adjustment, and tone mapping. Additionally, scene points may exhibit varying color details and the changing appearance due to directional lighting, such as distinct highlights and shadows. Intrinsic properties, such as material characteristics of the scene, remain constant.
To effectively model these variations, we explicitly decouple the appearance into three distinct components:
\textbf{Global Appearance Feature} (\(f_g \in \mathbb{R}^{n_g}\)): Captures information about the overall scene.
\textbf{Refined Appearance Feature} (\(f_r \in \mathbb{R}^{n_r}\)): Contains details beyond global features, specific to regional positions within the scene, such as high-frequency details and local highlights and shadows.
\textbf{Intrinsic Feature} (\(f_v \in \mathbb{R}^{n_v}\)): Represents the inherent properties of the scene points.

For a point \(v\) located at \(\mathbf{x}_i\) in 3D space, its dynamic appearance is characterized by these three components. The global and refined appearance features (\(f_g\) and \(f_r\)) are extracted from a reference image, while the intrinsic feature \(f_v\) is learned during the training process. This structured and explicit appearance decoupling effectively balances global scene information with local contextual details, while preserving intrinsic scene properties.

In our implementation, we utilize a voxel-based organization of Gaussians as described in Scaffold-GS \cite{lu2024scaffold}. Each anchor point \(v\) at the center of a voxel is associated with a scaling factor \(l_v \in \mathbb{R}^3\) and \(k\) learnable offsets \(O_v \in \mathbb{R}^{k \times 3}\), which collectively manage the \(k\) Gaussians within the voxel. To model global appearance changes consistently across the scene, we assign a uniform global appearance feature \(f_g\), extracted from a reference image, to all anchors. This extraction is performed by passing the feature map obtained from the UNet encoder through a global average pooling layer, followed by a trainable MLP \(MLP^G\) to obtain \(f_g\).

\subsection{Micro-macro Wavelet-based Sampling}
To enhance the accuracy of scene representation, we introduce a novel approach called Micro-Macro Wavelet-Based Sampling (MWS). This method refines the appearance features of each 3D Gaussian by capturing more detailed and diverse information. This approach aligns more closely with real-world scene variations by incorporating both fine-grained and broad-scale features. The MWS strategy consists of two main components:

\begin{figure}[t]    
  \centering        
  \subfloat[GS-W \cite{zhang2024gaussian-w}.]
  {
      \includegraphics[width=0.45\columnwidth]{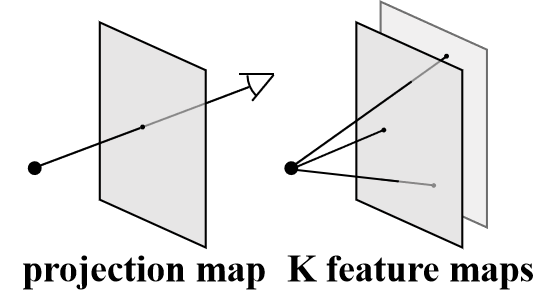}
      \label{fig:gs-w_sample}
  }
  \subfloat[Our MW-GS.]
  {
      \includegraphics[width=0.48\columnwidth]{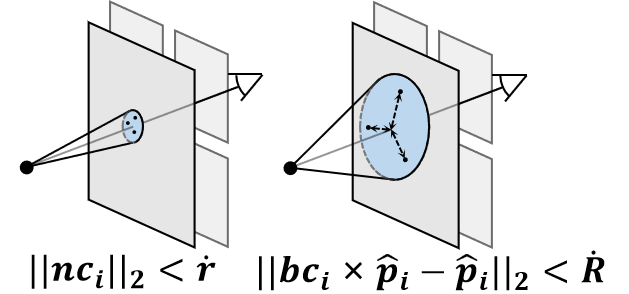}
      \label{fig:our_sample}
  }
  \caption{Sampling comparison between GS-W and the proposed MW-GS. Our approach integrates both narrow and broad conical frustums with wavelet-based sampling, allowing for a more comprehensive capture of features and resulting in enhanced accuracy.}    
  \label{fig:sampling}  
\end{figure}
\noindent\textbf{Micro-macro Projection (MP).}
In traditional MipMap techniques  \cite{williams1983pyramidal}, jitter sampling \cite{cook1986stochastic} is used to introduce random perturbations, enhancing the representation of texture details. Building on this concept, we propose an adaptive jitter projection method to achieve micro-projections. Rather than projecting each 3D point directly onto a fixed position on the 2D feature map, we project each point within a narrow conical frustum. This approach allows each Gaussian along a ray to capture similar yet distinct features, effectively reflecting the unique properties of each point in 3D space.

Fig. \ref{fig:gs-w_sample} presents a comparison between our method and GS-W. The GS-W method starts by directly projecting onto the \textit{projection feature map}, resulting in identical local appearance features for 3D points along the same ray. While GS-W attempts to mitigate this by adaptively sampling across multiple feature maps with varying positions, it lacks explicit control over targeting specific local regions. Consequently, the informative features within these regions are not fully utilized. 
In contrast, our micro projection method employs a narrow conical frustum to project 3D points onto the 2D feature map, with the cross-sectional radius parameterized by \( \dot{r} \). To further refine this process, we assign \( k_s \) learnable coordinate offsets \(\{nc_i\}_{k_s}\) to each 3D point, enabling adaptive sampling within the frustum. The features obtained from these \( k_s \) samples are then averaged to produce the refined feature \( f^n_r \). This approach ensures that feature sampling is diverse but still consistent, capturing rich fine-grained details while maintaining the coherence of rendered textures. By balancing variability and accuracy, our method significantly enhances representation performance.

Additionally, refined appearance features extend beyond fine details to encompass broader, long-range characteristics such as highlights. To capture these broader aspects, we also implement sampling through a broader conical frustum, as depicted on the right side of Fig. \ref{fig:our_sample}. Inspired by the principle that the size of point projection is inversely proportional to its distance from the camera, we parameterize the projection radius of the broad conical frustum as \( \dot{R} = \dot{R}_{max} / \| \mathbf{x}_i - \mathbf{x}_c \|_2 \), where \( \mathbf{x}_c \) represents the camera center. To avoid excessively small radius, we impose a minimum threshold \( \dot{R}_{min} \). By assigning \( k_s \) learnable coordinate scaling factors \(\{bc_i\}_{k_s}\) to each 3D point, we facilitate adaptive sampling within the broad conical frustum using the implementation \(bc_i \times \hat{p}_i\), where \(\hat{p}_i\) represents the projection center of the frustum. This design encourages a broader sampling range and enhances the representation of long-range features. The features derived from these \( k_s \) samples are averaged to generate the broad appearance feature \( f^b_r \).

\begin{table*}[t!]
\renewcommand\arraystretch{0.8}
\centering
\begin{tabular}{cccccccccc}
\toprule
\multirow{2}{*}{Method} & \multicolumn{3}{c}{\textit{Brandenburg Gate}} & \multicolumn{3}{c}{\textit{Sacre Coeur}} & \multicolumn{3}{c}{\textit{Trevi Fountain}} \\ 
\cmidrule(lr){2-4} \cmidrule(lr){5-7} \cmidrule(lr){8-10} 
& PSNR $\uparrow$ & SSIM $\uparrow$ & LPIPS $\downarrow$ & PSNR $\uparrow$ & SSIM $\uparrow$ & LPIPS $\downarrow$ & PSNR $\uparrow$ & SSIM $\uparrow$ & LPIPS $\downarrow$ \\
\midrule
NeRF-W & 24.17 & 0.890 & 0.167 & 19.20 & 0.807 & 0.191 & 18.97 & 0.698 & 0.265 \\
Ha-NeRF & 24.04 & 0.887 & 0.139 & 20.02 & 0.801 & 0.171 & 20.18 & 0.690 & 0.223 \\
CR-NeRF & 26.53 & 0.900 &  0.106 & 22.07 & 0.823 & 0.152  & 21.48 & 0.712 &  0.207 \\
WildGaussians &  27.77 &  0.927 & 0.133 &  22.56 &   0.859 & 0.177 & \underline{23.63} &  0.766 & 0.228 \\
GS-W & \underline{27.96} & \underline{0.931} & \underline{0.086} & \underline{23.24} & \underline{0.863} & \underline{0.130} &  22.91 & \underline{0.801} & \underline{0.156} \\
\midrule
Ours &  \textbf{29.37} &  \textbf{0.942} &  \textbf{0.052} &  \textbf{24.64} &  \textbf{0.897} &  \textbf{0.073} &  \textbf{24.07} &  \textbf{0.821} &  \textbf{0.120} \\
\bottomrule
\end{tabular}
\caption{Quantitative results on three datasets. \textbf{Bold} and \underline{underlined} values correspond to the best and the second-best value, respectively. Our method outperforms the previous methods across all datasets on PSNR , SSIM, and LPIPS.}
\label{tab:quantitative}
\end{table*}

\noindent\textbf{Wavelet-based Sampling (WS).}
In unconstrained image collections, camera parameters can vary significantly, making it challenging to handle large-scale variations across different viewpoints with fixed-resolution sampling alone. To address this, we introduce a Wavelet-based Sampling technique that enhances the capture of high-frequency and multi-scale information. Specifically, we apply the Discrete Wavelet Transform (DWT) to the feature map \( \mathbf{F}^{MAP} \) generated by the UNet, producing a series of feature maps. The DWT decomposes the feature map into four frequency bands while simultaneously reducing its resolution, effectively preserving the spatial information and enabling multi-scale sampling that captures diverse frequency information.

The process begins by segmenting the feature map \( \mathbf{F}^{MAP} \) into \(2M+2\) smaller feature maps \(\{\mathbf{F}^1, \cdots, \mathbf{F}^{2M+2}\}\), where each \(\mathbf{F}^i \in \mathbb{R} ^{\frac{n_r}{2M+2}\times H^\mathbf{F}\times W^\mathbf{F}}\). Here, \(M\) represents the maximum number of downsampling operations or the highest level of Discrete Wavelet Transform (DWT), which serves as a critical hyper-parameter. The dimensions \( H^\mathbf{F} \) and \( W^\mathbf{F} \) correspond to the height and width of the smaller feature maps, respectively.

During the \(m\)-th downsampling stage, an \(m\)-level DWT is applied to the \((2m+1)\)-th and \((2m+2)\)-th feature maps, resulting in \( 4^m \) sub-feature maps as per Eq. (\ref{eq:wavelet}). These sub-feature maps are then subjected to bilinear interpolation sampling through narrow and broad conical frustums, employing the Micro-macro Projection technique, which yields the feature sets \( \{f^n_{r,m,j}\}_{4^m} \) and \( \{f^b_{r,m,j}\}_{4^m} \), respectively.

Next, the refined features \( f^n_{r,m} \) and \( f^b_{r,m} \) for each downsampling level are computed by applying learnable weight parameters to the sampled features:
\begin{equation}
    f^n_{r,m} = \sum_{j=1}^{4^m} \omega^n_{m,j} \cdot f^n_{r,m,j}, \;
    f^b_{r,m}= \sum_{j=1}^{4^m} \omega^b_{m,j} \cdot f^b_{r,m,j},
\end{equation}
where \( \omega^n_{m,j} \) and \( \omega^b_{m,j} \) are the learnable weights for the \((2m+1)\)-th and \((2m+2)\)-th feature maps, respectively.
Finally, the refined appearance features corresponding to each anchor are obtained by concatenating the features from all scales:
\begin{equation}
    f_r = f^n_{r,0} \oplus f^b_{r,0} \oplus\cdots\oplus f^n_{r,M} \oplus f^b_{r,M}.
\end{equation}

The combination of Micro-macro Projection and Wavelet-based Sampling supplements scene representation with multi-scale and high-frequency features. It accurately captures refined appearance variations, providing a comprehensive understanding of scene structures at multiple scales.

\subsection{Hierarchical Residual Fusion Network}
\label{sec:hrfn}
After obtaining the global appearance, refined appearance, and intrinsic features, it is essential to effectively combine these with the position and view direction to generate the final \( k \) Gaussian colors corresponding to each anchor. Given that these features exist in different high-dimensional spaces, simple concatenation would not achieve an effective fusion. To address this, we propose a Hierarchical Residual Fusion Network (HRFN), which consists of four Multi-Layer Perceptrons (MLPs), denoted as \( \mathcal{M}^H = \{ \mathcal{M}^H_1, \mathcal{M}^H_2, \mathcal{M}^H_3, \mathcal{M}^H_4 \} \).

The inputs to the network include the position of the anchor center \(\mathbf{x}_i\), global appearance feature \(f_g\), refined appearance feature \(f_r\), intrinsic features \(f_v\), and the direction vector \(\vec{\mathbf{d}}_{ic} = \frac{\mathbf{x}_i - \mathbf{x}_c}{||\mathbf{x}_i - \mathbf{x}_c||_2}\). The network processes these inputs to infer the output colors \(\{\hat{c}_k\}\) for the \( k \) Gaussians as follows:

\begin{equation}
\begin{aligned}
    &Emb = \mathcal{M}^H_1(\gamma(\mathbf{x}_i)\oplus f_v\oplus f_r\oplus f_g) \oplus \omega_r f_r, \\
    &\{\hat{c}_k\} = \mathcal{M}^H_4\left(\mathcal{M}^H_3\left(\mathcal{M}^H_2\left(Emb \oplus \omega_v f_v\right)\right) \oplus \vec{\mathbf{d}}_{ic}\right),
\end{aligned}
\end{equation}
where \(\oplus\) denotes concatenation, \(\gamma(\cdot)\) is the positional encoding function, and \(\omega_r\) and \(\omega_v\) are learnable adaptive weights. This hierarchical structure enables the network to effectively fuse features from different levels, ensuring a comprehensive integration of global, refined, and intrinsic information. This design improves the network's ability to accurately predict the final Gaussian colors by capturing the complex interactions among the various input features.

\subsection{Handling Transient Objects and Optimization}
Transient objects with view inconsistencies are widespread in unconstrained photo collections, posing significant challenges for 3DGS, which aim to fit each view accurately.
To address this issue, we reuse the UNet encoder that extracts appearance features, and adding a separate decoder to predict the visibility map $vm$. Once the $vm$ is obtained, it is applied to the encoded intermediate features of the appearance feature extraction network.
Additionally, a visibility map regularization is applied to prevent meaningful pixels from being masked out. The overall loss function between the rendered image \( I_r \) and the reference image \( I_{gt} \), incorporating the visibility map, can be expressed as follows:
\begin{equation}
\begin{aligned}
    \mathcal{L} &=\lambda_{SSIM}\mathcal{L}_{SSIM}(vm\odot I_r, vm\odot I_{gt}) \\
    &+ \lambda_1\mathcal{L}_1(vm\odot I_r, vm\odot I_{gt}) + \lambda_{vm}\mathcal{L}_2(vm, 1),
\end{aligned}
\end{equation}
where \(\odot\) denotes the Hadamard product.

\begin{figure*}[t]
\centering
\includegraphics[width=0.9\textwidth, height=0.45\textwidth]{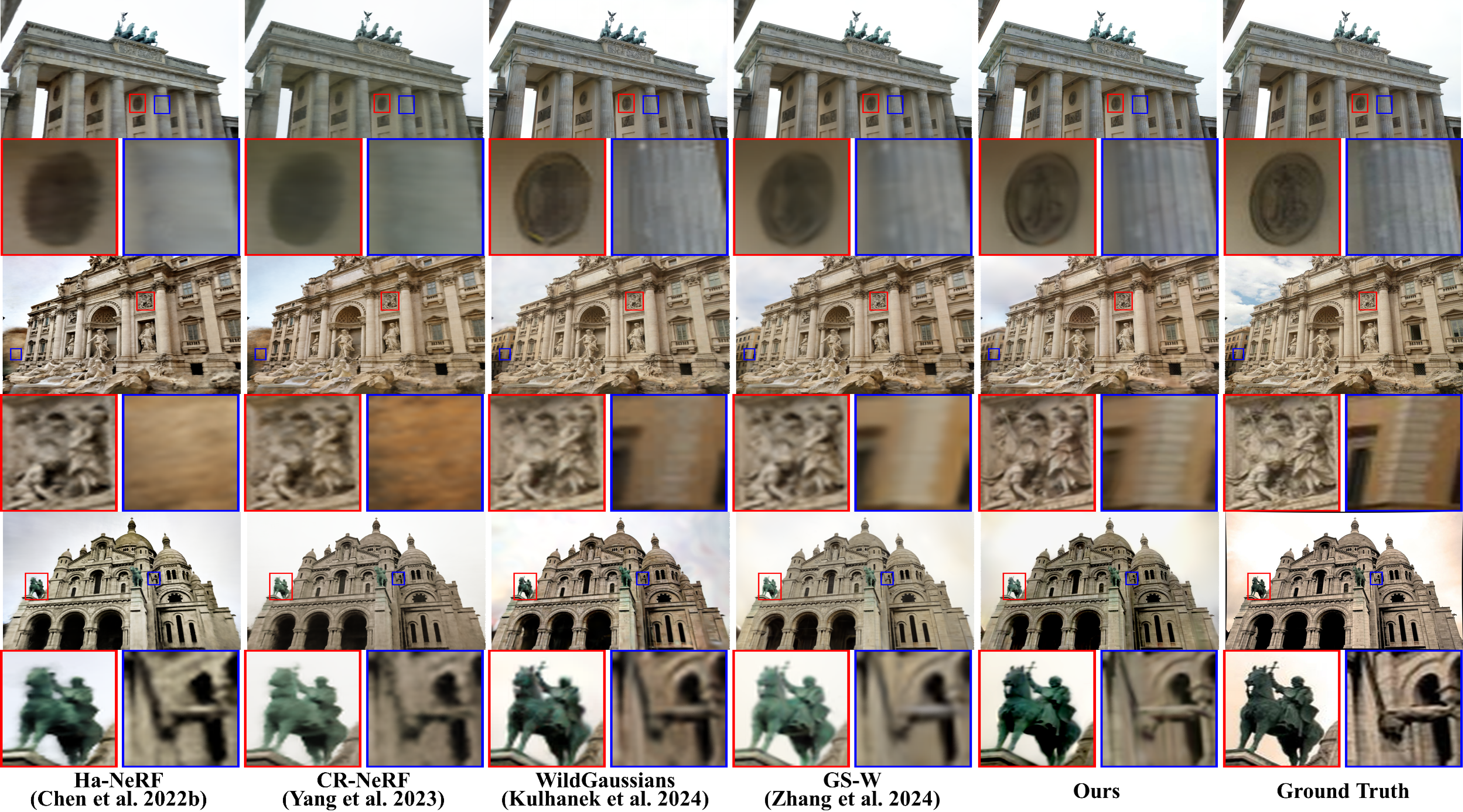}
\caption{Qualitative comparison on three datasets. Red and blue crops emphasize that MW-GS can recover finer details.}
\label{fig:qualitative}
\end{figure*}

\begin{table*}[t]
\renewcommand\arraystretch{0.8}
\centering
\begin{tabular}{cccccccccc}
\toprule
\multirow{2}{*}{Method} & \multicolumn{3}{c}{\textit{Brandenburg Gate}} & \multicolumn{3}{c}{\textit{Sacre Coeur}} & \multicolumn{3}{c}{\textit{Trevi Fountain}} \\ 
\cmidrule(lr){2-4} \cmidrule(lr){5-7} \cmidrule(lr){8-10} 
& PSNR $\uparrow$ & SSIM $\uparrow$ & LPIPS $\downarrow$ & PSNR $\uparrow$ & SSIM $\uparrow$ & LPIPS $\downarrow$ & PSNR $\uparrow$ & SSIM $\uparrow$ & LPIPS $\downarrow$ \\
\midrule
w/o WS & \underline{28.88}  & \underline{0.939}  & \underline{0.055} & 24.35  & 0.895 & 0.075 & 23.38  & 0.810 & 0.127 \\
only micro & 28.49 &  0.939 & 0.059 & 23.83  & 0.888  & 0.081 & 23.47  & 0.809  & 0.129 \\
w/o HRFN & 28.69  & 0.936  & 0.057 & 24.45 & \underline{0.896} & \underline{0.074} & \underline{23.68} & 0.813 & 0.125\\
w/o vm & 28.66 &  0.934 &  0.057 &  \textbf{24.83} & 0.893 & 0.074 & 23.65 & \underline{0.815} & \underline{0.122} \\
\midrule
Full model &  \textbf{29.37} &  \textbf{0.942} &  \textbf{0.052} & \underline{24.64} &  \textbf{0.897} &  \textbf{0.073} &  \textbf{24.07} &  \textbf{0.821} &  \textbf{0.120} \\
\bottomrule
\end{tabular}
\caption{Ablation studies on three datasets. \textbf{Bold} and \underline{underlined} values correspond to the best and the second-best value.}
\label{tab:ablation}
\end{table*}

\section{Experiment}
\subsection{Datasets and Metrics}
Following previous works \cite{chen2022hallucinated, zhang2024gaussian-w}, we evaluate different methods on three datasets: \textit{Brandenburg Gate}, \textit{Sacre Coeur}, and \textit{Trevi Fountain}, with all images downsampled by a factor of 2 during both training and evaluation. We use PSNR, SSIM \cite{wang2004image}, and LPIPS \cite{zhang2018unreasonable} as metrics to assess the performance of our method. We compare it against NeRF-W \cite{martin2021nerf-w}, Ha-NeRF \cite{chen2022hallucinated}, CR-NeRF \cite{yang2023cross}, WildGaussians \cite{kulhanek2024wildgaussians} and GS-W \cite{zhang2024gaussian-w}.

\subsection{Comparison Experiments}
\noindent\textbf{Quantitative comparison.} 
The quantitative results presented in Tab. \ref{tab:quantitative} highlight the efficacy of our MW-GS method. NeRF-W introduces global appearance and transient embeddings, providing a basic level of scene reconstruction. Ha-NeRF and CR-NeRF build on this foundation, achieving improvements, but they still fall short in adequately capturing the local contextual information of various scene points. Similarly, WildGaussians exhibit limitations in appearance modeling due to their dependence on global appearance embeddings. GS-W addresses this by employing adaptive sampling of local features, leading to enhanced focus on detailed information, as reflected in consistently improved metrics.
Our MW-GS method advances beyond GS-W by seamlessly integrating long-range contextual information with detailed localized features within the Gaussian representations. Furthermore, it enriches multi-scale information and captures high-frequency details more effectively. This approach yields significant improvements, outperforming GS-W by 1.41 dB, 1.4 dB, and 1.16 dB in PSNR across the three evaluated scenes.
\begin{figure*}[ht!]    
  \centering        
  \subfloat[Appearance transfer using reference images inside and outside the dataset.]
  {
      \includegraphics[width=0.55\textwidth, height=0.30\textwidth]{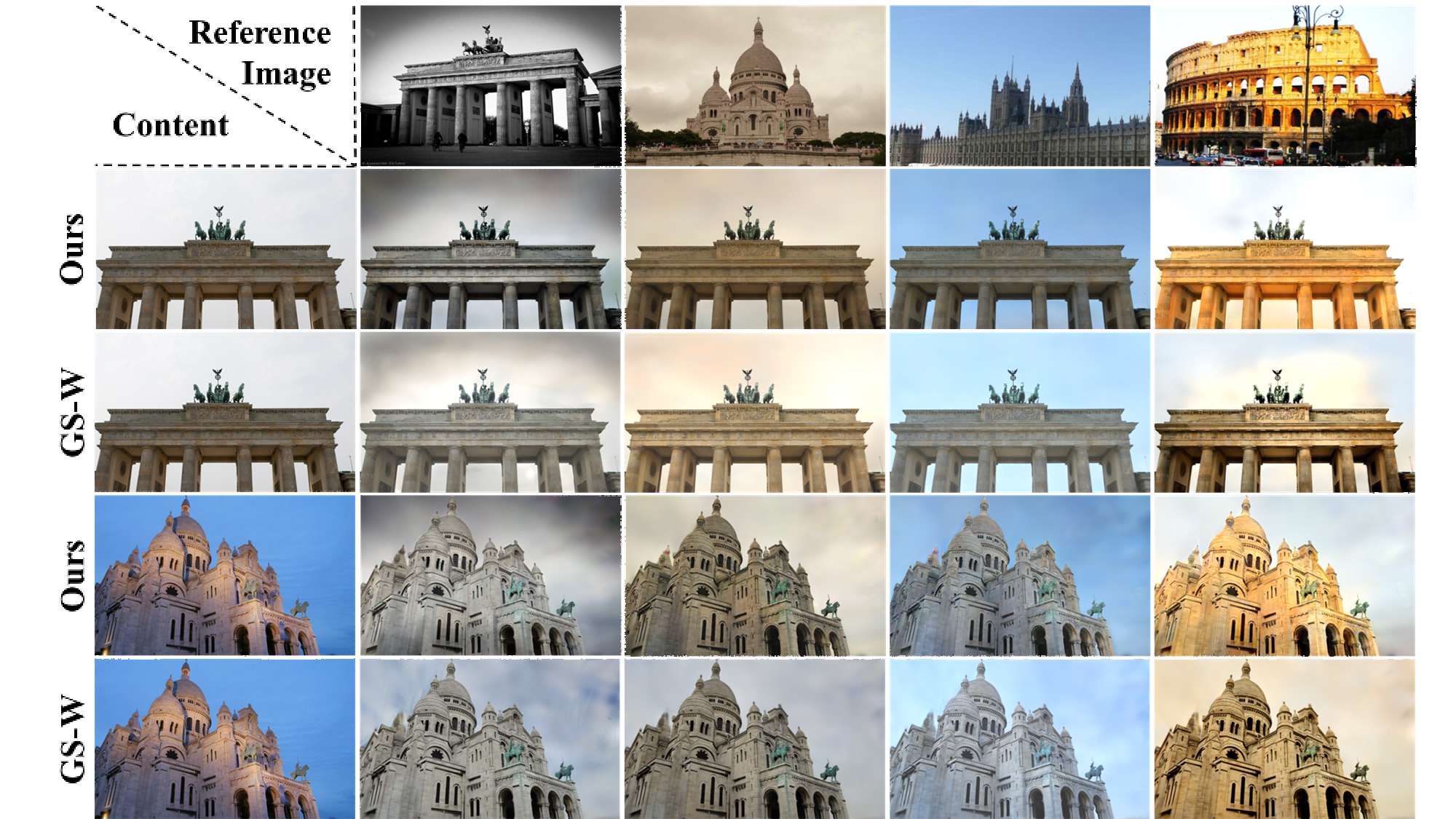}
      \label{fig:transfer}
  }
  \subfloat[Appearance tuning by global and local features.]
  {
      \includegraphics[width=0.37\textwidth, height=0.30\textwidth]{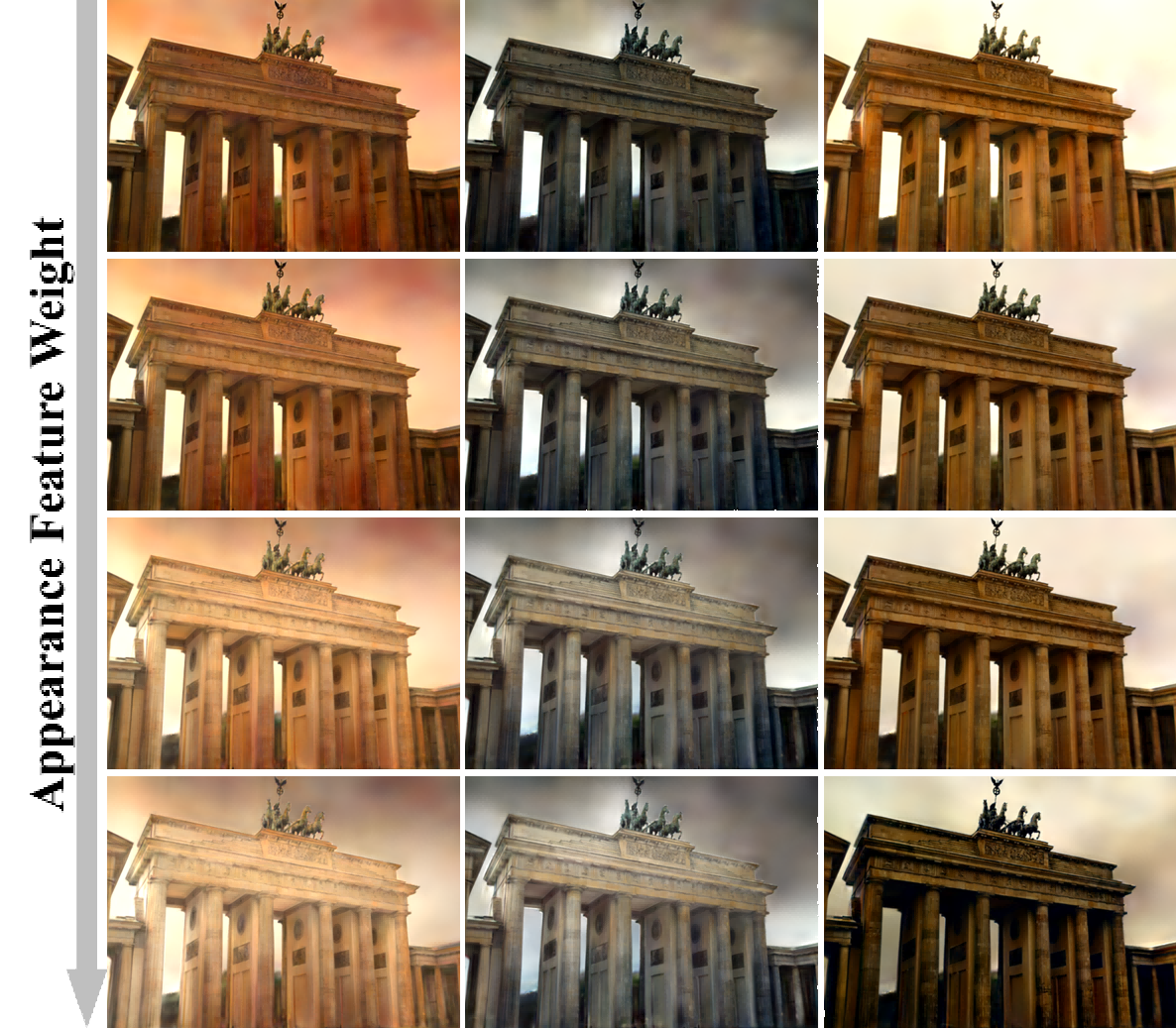}
      \label{fig:app_tuning}
  }
  \caption{Qualitative results of appearance transfer.}    
  \label{fig:appearance}  
\end{figure*}

\begin{figure}[t]    
\centering
\includegraphics[width=0.98\columnwidth]{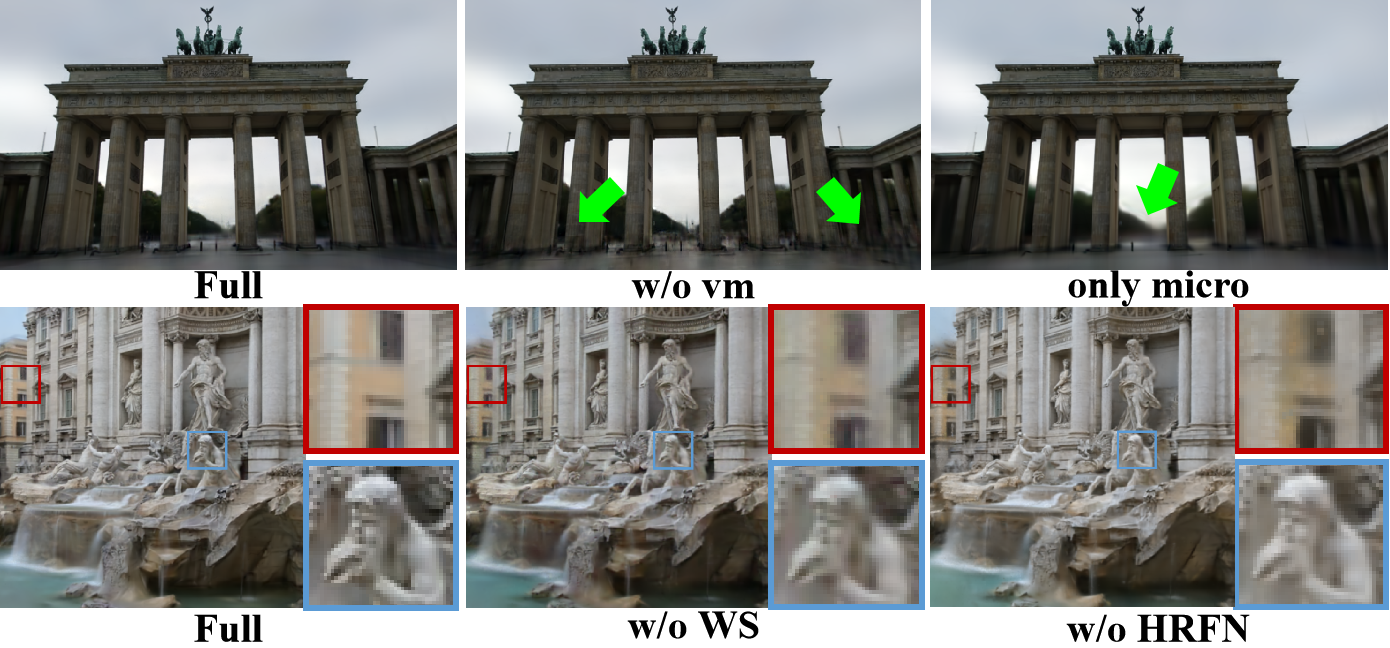}
\caption{Ablation studies by visualization.}    
\label{fig:ablation}
\end{figure}

\noindent\textbf{Qualitative comparison.} 
The qualitative results in Fig. \ref{fig:qualitative} clearly demonstrate the strengths of our approach. Existing methods often struggle to capture intricate scene details and complex textures accurately. In contrast, our method excels by utilizing micro-macro wavelet-based sampling to enhance feature extraction, effectively integrating frequency domain and multi-scale information. Furthermore, the hierarchical fusion of structured features allows for a more precise recovery of appearance details and clear structural representation. This results in superior visual fidelity and accuracy in the reconstructed scenes. For example, our method successfully captures finer details and more accurate colors in the reliefs of the Trevi Fountain and the bronze statues at Sacre Coeur, outperforming current techniques.

\subsection{Component Analysis}
\noindent\textbf{Ablation Study.}
The ablation study results achieved on three datasets are summarized in Tab. \ref{tab:ablation} and qualitative outcomes are illustrated in Fig. \ref{fig:ablation}. Key findings include:
1) MP significantly enhances the diversity of refined appearance sampling.
As demonstrated in Fig. \ref{fig:ablation}, relying solely on micro projection, without full MP and WS, results in noticeable blurring of distant objects.
2) WS further refines attention to high-frequency and multi-scale information.
Without WS, there is a marked loss of detail, as evidenced by the blurring of the Trevi Fountain sculptures and a corresponding 0.69 dB decrease in PSNR.
3) HRFN
facilitates a comprehensive fusion of diverse information compared to simple concatenation. This results in a 0.68 dB increase in PSNR on the \textit{Brandenburg Gate}.
4) The removal of the visibility map introduces significant artifacts in the rendered images, as seen in Fig. \ref{fig:ablation}, underscoring its importance in maintaining visual integrity.

\noindent\textbf{Analysis of Appearance Transfer.}  
Our method demonstrates strong support for appearance transfer in 3D scenes, showing its advanced appearance modeling capabilities. As illustrated in Fig. \ref{fig:transfer}, a qualitative comparison between GS-W and our approach reveals that our method not only transfers both foreground and background elements to novel views but also preserves the intricate details of the scene, rather than merely replicating the overall scene tone. This capability underscores the accuracy and robustness of our appearance modeling.
Furthermore, our approach allows for the adjustment of dynamic appearance features during rendering by modulating their corresponding weights. As depicted in Fig. \ref{fig:app_tuning}, altering these weights can effectively adjust the overall exposure of the rendered image. Additionally, the variation in intrinsic features across different regions results in corresponding changes in local highlights, emphasizing the importance and effectiveness of disentangling and capturing intrinsic features.

\begin{figure}[t]    
  \centering        
  \subfloat[Sampling attention.]
  {
      \includegraphics[width=0.34\columnwidth]{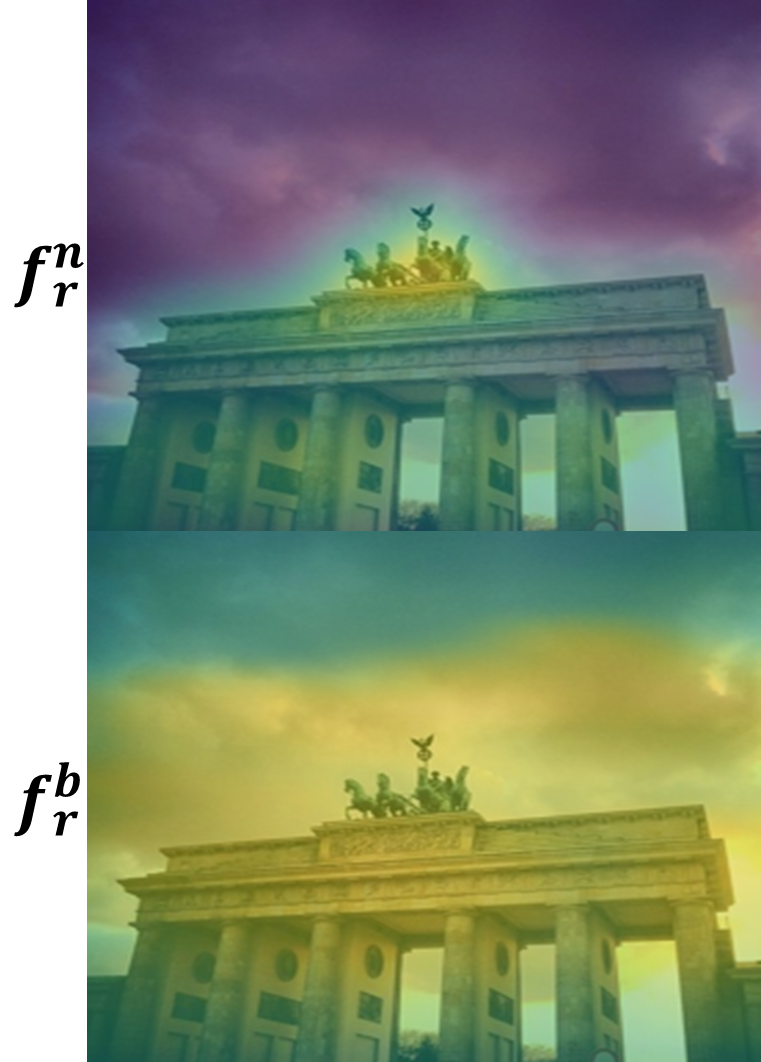}
      \label{fig:feat_attention}
  }
  \subfloat[Sampling features rendering result.]
  {
      \includegraphics[width=0.63\columnwidth]{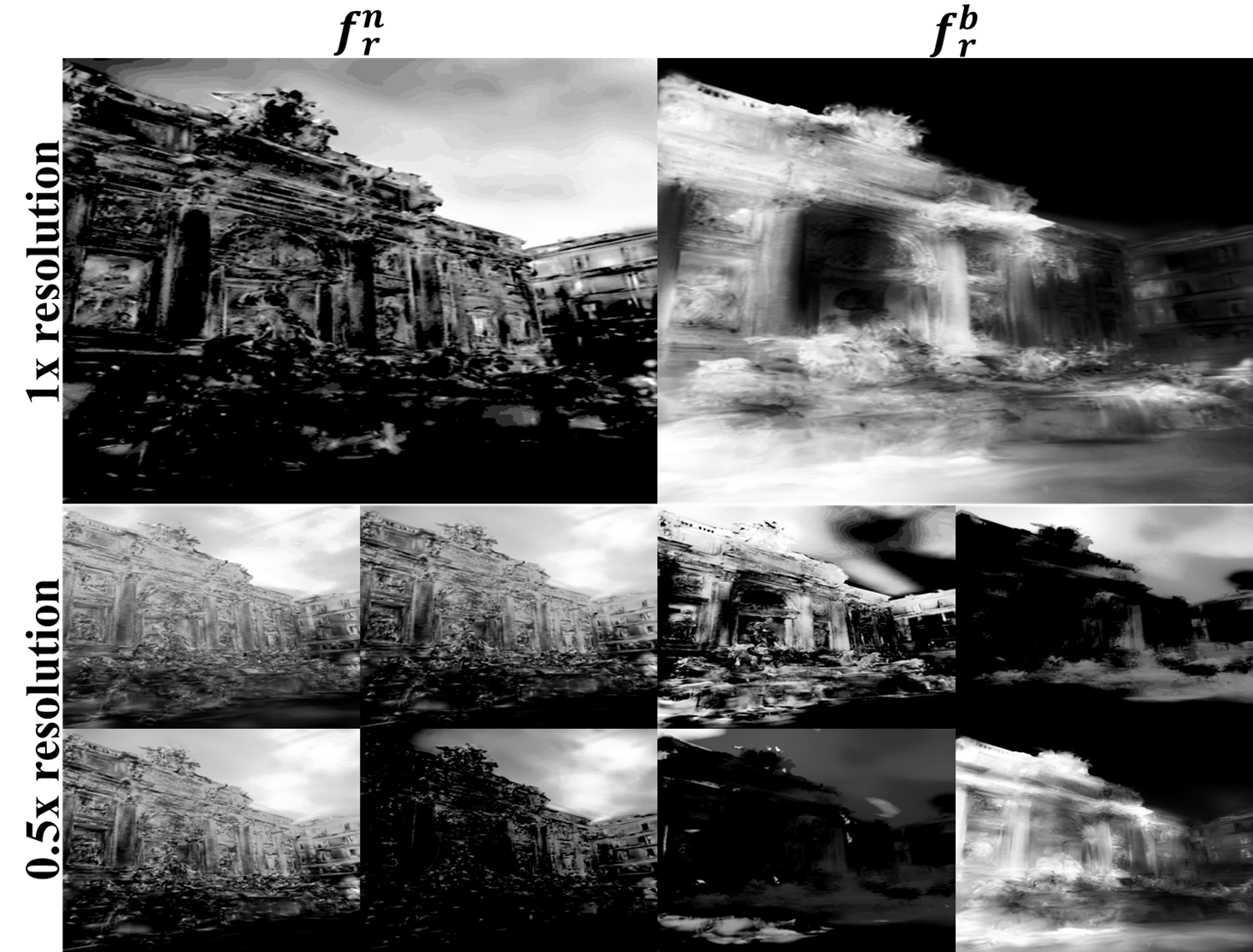}
      \label{fig:feat_sampling}
  }
  \caption{Visualization of sampling analysis.}  
  \label{fig:sampling-analysis}  
\end{figure}

\noindent\textbf{Analysis of Sampling.} 
We project the sampling positions onto corresponding camera images to generate an attention map, where areas with dense sampling points indicate high attention. As shown in Fig. \ref{fig:feat_attention}, our sampling approach effectively captures fine-grained local details and long-range information by utilizing narrow and broad conical frustum projections, respectively.
We further visualize the features of interest by examining the refined narrow features \( f^n_r \) and broad features \( f^b_r \) at different resolutions, as depicted in Fig. \ref{fig:feat_sampling}. The \( f^n_r \) features, focused on high-resolution details, effectively capture local texture intricacies, while the 0.5x resolution features, processed through DWT, attend to varying details across different frequency bands. In contrast, the \( f^b_r \) features primarily target low-texture regions such as water surfaces or highlights, which correspond to long-range features. 
The combination of both enables our refined appearance feature to effectively model dynamic appearances.

\section{Conclusion}
In this paper, we present MW-GS, a novel approach for handling varying appearances in unconstrained photo collections for novel view synthesis. By decomposing appearance features into global, refined, and intrinsic components and employing Micro-macro Wavelet-based Sampling, our method captures precise and diverse appearance features while preserving multi-scale and frequency domain details. 
Extensive experiments show that MW-GS outperforms existing methods, setting a new standard in rendering quality and adaptability to dynamic scenes.

\section{Acknowledgments}
This work is partly supported by the National Key R\&D Program of China (2022ZD0161902),  the National Natural Science Foundation of China (No. 62202031, 62176012), the Fundamental Research Funds for the Central Universities, the Research Program of State Key Laboratory of Critical Software Environment, and the Fundamental Research Funds for the Central Universities. 

\bibliography{aaai25}

\begin{thebibliography}{53}
\providecommand{\natexlab}[1]{#1}

\bibitem[{Barron et~al.(2022)Barron, Mildenhall, Verbin, Srinivasan, and
  Hedman}]{barron2022mip}
Barron, J.~T.; Mildenhall, B.; Verbin, D.; Srinivasan, P.~P.; and Hedman, P.
  2022.
\newblock Mip-nerf 360: Unbounded anti-aliased neural radiance fields.
\newblock In \emph{Proceedings of the IEEE/CVF conference on computer vision
  and pattern recognition}, 5470--5479.

\bibitem[{Barron et~al.(2023)Barron, Mildenhall, Verbin, Srinivasan, and
  Hedman}]{barron2023zip}
Barron, J.~T.; Mildenhall, B.; Verbin, D.; Srinivasan, P.~P.; and Hedman, P.
  2023.
\newblock Zip-nerf: Anti-aliased grid-based neural radiance fields.
\newblock In \emph{Proceedings of the IEEE/CVF International Conference on
  Computer Vision}, 19697--19705.

\bibitem[{Chen et~al.(2022{\natexlab{a}})Chen, Xu, Geiger, Yu, and
  Su}]{chen2022tensorf}
Chen, A.; Xu, Z.; Geiger, A.; Yu, J.; and Su, H. 2022{\natexlab{a}}.
\newblock Tensorf: Tensorial radiance fields.
\newblock In \emph{European conference on computer vision}, 333--350. Springer.

\bibitem[{Chen et~al.(2021)Chen, Yao, Chen, Ding, Li, and Ji}]{chen2021local}
Chen, S.; Yao, T.; Chen, Y.; Ding, S.; Li, J.; and Ji, R. 2021.
\newblock Local relation learning for face forgery detection.
\newblock In \emph{Proceedings of the AAAI Conference on Artificial
  Intelligence}, volume~35, 1081--1088.

\bibitem[{Chen et~al.(2022{\natexlab{b}})Chen, Zhang, Li, Chen, Feng, Wang, and
  Wang}]{chen2022hallucinated}
Chen, X.; Zhang, Q.; Li, X.; Chen, Y.; Feng, Y.; Wang, X.; and Wang, J.
  2022{\natexlab{b}}.
\newblock Hallucinated neural radiance fields in the wild.
\newblock In \emph{Proceedings of the IEEE/CVF Conference on Computer Vision
  and Pattern Recognition}, 12943--12952.

\bibitem[{Chen et~al.(2024{\natexlab{a}})Chen, Chen, Zhang, Wang, Yang, Wang,
  Cai, Yang, Liu, and Lin}]{chen2024gaussianeditor}
Chen, Y.; Chen, Z.; Zhang, C.; Wang, F.; Yang, X.; Wang, Y.; Cai, Z.; Yang, L.;
  Liu, H.; and Lin, G. 2024{\natexlab{a}}.
\newblock Gaussianeditor: Swift and controllable 3d editing with gaussian
  splatting.
\newblock In \emph{Proceedings of the IEEE/CVF Conference on Computer Vision
  and Pattern Recognition}, 21476--21485.

\bibitem[{Chen et~al.(2023)Chen, Funkhouser, Hedman, and
  Tagliasacchi}]{chen2023mobilenerf}
Chen, Z.; Funkhouser, T.; Hedman, P.; and Tagliasacchi, A. 2023.
\newblock Mobilenerf: Exploiting the polygon rasterization pipeline for
  efficient neural field rendering on mobile architectures.
\newblock In \emph{Proceedings of the IEEE/CVF Conference on Computer Vision
  and Pattern Recognition}, 16569--16578.

\bibitem[{Chen et~al.(2024{\natexlab{b}})Chen, Wang, Wang, and
  Liu}]{chen2024text}
Chen, Z.; Wang, F.; Wang, Y.; and Liu, H. 2024{\natexlab{b}}.
\newblock Text-to-3d using gaussian splatting.
\newblock In \emph{Proceedings of the IEEE/CVF Conference on Computer Vision
  and Pattern Recognition}, 21401--21412.

\bibitem[{Chung et~al.(2023)Chung, Lee, Nam, Lee, and
  Lee}]{chung2023luciddreamer}
Chung, J.; Lee, S.; Nam, H.; Lee, J.; and Lee, K.~M. 2023.
\newblock Luciddreamer: Domain-free generation of 3d gaussian splatting scenes.
\newblock \emph{arXiv preprint arXiv:2311.13384}.

\bibitem[{Cook(1986)}]{cook1986stochastic}
Cook, R.~L. 1986.
\newblock Stochastic sampling in computer graphics.
\newblock \emph{ACM Transactions on Graphics (TOG)}, 5(1): 51--72.

\bibitem[{Dahmani et~al.(2024)Dahmani, Bennehar, Piasco, Roldao, and
  Tsishkou}]{dahmani2024swag}
Dahmani, H.; Bennehar, M.; Piasco, N.; Roldao, L.; and Tsishkou, D. 2024.
\newblock SWAG: Splatting in the Wild images with Appearance-conditioned
  Gaussians.
\newblock \emph{arXiv preprint arXiv:2403.10427}.

\bibitem[{Daubechies(1992)}]{daubechies1992ten}
Daubechies, I. 1992.
\newblock \emph{Ten lectures on wavelets}.
\newblock Houthalen-Helchteren: SIAM.

\bibitem[{Duan et~al.(2017)Duan, Liu, Jiao, Zhao, and Zhang}]{duan2017sar}
Duan, Y.; Liu, F.; Jiao, L.; Zhao, P.; and Zhang, L. 2017.
\newblock SAR image segmentation based on convolutional-wavelet neural network
  and Markov random field.
\newblock \emph{Pattern Recognition}, 64: 255--267.

\bibitem[{Groueix et~al.(2018)Groueix, Fisher, Kim, Russell, and
  Aubry}]{groueix2018papier}
Groueix, T.; Fisher, M.; Kim, V.~G.; Russell, B.~C.; and Aubry, M. 2018.
\newblock A papier-m{\^a}ch{\'e} approach to learning 3d surface generation.
\newblock In \emph{Proceedings of the IEEE conference on computer vision and
  pattern recognition}, 216--224.

\bibitem[{Hu et~al.(2023)Hu, Wang, Ma, Yang, Gao, Liu, and Ma}]{hu2023tri}
Hu, W.; Wang, Y.; Ma, L.; Yang, B.; Gao, L.; Liu, X.; and Ma, Y. 2023.
\newblock Tri-miprf: Tri-mip representation for efficient anti-aliasing neural
  radiance fields.
\newblock In \emph{Proceedings of the IEEE/CVF International Conference on
  Computer Vision}, 19774--19783.

\bibitem[{Huang et~al.(2024)Huang, Yu, Chen, Geiger, and Gao}]{huang20242d}
Huang, B.; Yu, Z.; Chen, A.; Geiger, A.; and Gao, S. 2024.
\newblock 2d gaussian splatting for geometrically accurate radiance fields.
\newblock In \emph{ACM SIGGRAPH 2024 Conference Papers}, 1--11.

\bibitem[{Kanazawa et~al.(2018)Kanazawa, Tulsiani, Efros, and
  Malik}]{kanazawa2018learning}
Kanazawa, A.; Tulsiani, S.; Efros, A.~A.; and Malik, J. 2018.
\newblock Learning category-specific mesh reconstruction from image
  collections.
\newblock In \emph{Proceedings of the European Conference on Computer Vision
  (ECCV)}, 371--386.

\bibitem[{Kato, Ushiku, and Harada(2018)}]{kato2018neural}
Kato, H.; Ushiku, Y.; and Harada, T. 2018.
\newblock Neural 3d mesh renderer.
\newblock In \emph{Proceedings of the IEEE conference on computer vision and
  pattern recognition}, 3907--3916.

\bibitem[{Kerbl et~al.(2023)Kerbl, Kopanas, Leimkuehler, and
  Drettakis}]{kerbl20233d}
Kerbl, B.; Kopanas, G.; Leimkuehler, T.; and Drettakis, G. 2023.
\newblock 3D Gaussian Splatting for Real-Time Radiance Field Rendering.
\newblock \emph{ACM Transactions on Graphics (TOG)}, 42(4): 1--14.

\bibitem[{Kulhanek et~al.(2024)Kulhanek, Peng, Kukelova, Pollefeys, and
  Sattler}]{kulhanek2024wildgaussians}
Kulhanek, J.; Peng, S.; Kukelova, Z.; Pollefeys, M.; and Sattler, T. 2024.
\newblock WildGaussians: 3D Gaussian Splatting in the Wild.
\newblock \emph{arXiv preprint arXiv:2407.08447}.

\bibitem[{Lee et~al.(2024)Lee, Rho, Sun, Ko, and Park}]{lee2024compact}
Lee, J.~C.; Rho, D.; Sun, X.; Ko, J.~H.; and Park, E. 2024.
\newblock Compact 3d gaussian representation for radiance field.
\newblock In \emph{Proceedings of the IEEE/CVF Conference on Computer Vision
  and Pattern Recognition}, 21719--21728.

\bibitem[{Li et~al.(2022)Li, Xie, Yu, and Zhang}]{li2022wavelet}
Li, J.; Xie, H.; Yu, L.; and Zhang, Y. 2022.
\newblock Wavelet-enhanced Weakly Supervised Local Feature Learning for Face
  Forgery Detection.
\newblock In \emph{Proceedings of the 30th ACM International Conference on
  Multimedia}, 1299--1308.

\bibitem[{Lu et~al.(2024)Lu, Yu, Xu, Xiangli, Wang, Lin, and
  Dai}]{lu2024scaffold}
Lu, T.; Yu, M.; Xu, L.; Xiangli, Y.; Wang, L.; Lin, D.; and Dai, B. 2024.
\newblock Scaffold-gs: Structured 3d gaussians for view-adaptive rendering.
\newblock In \emph{Proceedings of the IEEE/CVF Conference on Computer Vision
  and Pattern Recognition}, 20654--20664.

\bibitem[{Martin-Brualla et~al.(2021)Martin-Brualla, Radwan, Sajjadi, Barron,
  Dosovitskiy, and Duckworth}]{martin2021nerf-w}
Martin-Brualla, R.; Radwan, N.; Sajjadi, M.~S.; Barron, J.~T.; Dosovitskiy, A.;
  and Duckworth, D. 2021.
\newblock Nerf in the wild: Neural radiance fields for unconstrained photo
  collections.
\newblock In \emph{Proceedings of the IEEE/CVF conference on computer vision
  and pattern recognition}, 7210--7219.

\bibitem[{Mildenhall et~al.(2021)Mildenhall, Srinivasan, Tancik, Barron,
  Ramamoorthi, and Ng}]{mildenhall2021nerf}
Mildenhall, B.; Srinivasan, P.~P.; Tancik, M.; Barron, J.~T.; Ramamoorthi, R.;
  and Ng, R. 2021.
\newblock Nerf: Representing scenes as neural radiance fields for view
  synthesis.
\newblock \emph{Communications of the ACM}, 65(1): 99--106.

\bibitem[{M{\"u}ller et~al.(2022)M{\"u}ller, Evans, Schied, and
  Keller}]{muller2022instant}
M{\"u}ller, T.; Evans, A.; Schied, C.; and Keller, A. 2022.
\newblock Instant neural graphics primitives with a multiresolution hash
  encoding.
\newblock \emph{ACM transactions on graphics (TOG)}, 41(4): 1--15.

\bibitem[{Qi et~al.(2017{\natexlab{a}})Qi, Su, Mo, and Guibas}]{qi2017pointnet}
Qi, C.~R.; Su, H.; Mo, K.; and Guibas, L.~J. 2017{\natexlab{a}}.
\newblock Pointnet: Deep learning on point sets for 3d classification and
  segmentation.
\newblock In \emph{Proceedings of the IEEE conference on computer vision and
  pattern recognition}, 652--660.

\bibitem[{Qi et~al.(2017{\natexlab{b}})Qi, Yi, Su, and
  Guibas}]{qi2017pointnet++}
Qi, C.~R.; Yi, L.; Su, H.; and Guibas, L.~J. 2017{\natexlab{b}}.
\newblock Pointnet++: Deep hierarchical feature learning on point sets in a
  metric space.
\newblock \emph{Advances in neural information processing systems}, 30.

\bibitem[{Qin et~al.(2024)Qin, Li, Zhou, Wang, and Pfister}]{qin2024langsplat}
Qin, M.; Li, W.; Zhou, J.; Wang, H.; and Pfister, H. 2024.
\newblock Langsplat: 3d language gaussian splatting.
\newblock In \emph{Proceedings of the IEEE/CVF Conference on Computer Vision
  and Pattern Recognition}, 20051--20060.

\bibitem[{Reiser et~al.(2023)Reiser, Szeliski, Verbin, Srinivasan, Mildenhall,
  Geiger, Barron, and Hedman}]{reiser2023merf}
Reiser, C.; Szeliski, R.; Verbin, D.; Srinivasan, P.; Mildenhall, B.; Geiger,
  A.; Barron, J.; and Hedman, P. 2023.
\newblock Merf: Memory-efficient radiance fields for real-time view synthesis
  in unbounded scenes.
\newblock \emph{ACM Transactions on Graphics (TOG)}, 42(4): 1--12.

\bibitem[{Schwarz et~al.(2022)Schwarz, Sauer, Niemeyer, Liao, and
  Geiger}]{schwarz2022voxgraf}
Schwarz, K.; Sauer, A.; Niemeyer, M.; Liao, Y.; and Geiger, A. 2022.
\newblock Voxgraf: Fast 3d-aware image synthesis with sparse voxel grids.
\newblock \emph{Advances in Neural Information Processing Systems}, 35:
  33999--34011.

\bibitem[{Shi et~al.(2024)Shi, Wang, Duan, and Guan}]{shi2024language}
Shi, J.-C.; Wang, M.; Duan, H.-B.; and Guan, S.-H. 2024.
\newblock Language embedded 3d gaussians for open-vocabulary scene
  understanding.
\newblock In \emph{Proceedings of the IEEE/CVF Conference on Computer Vision
  and Pattern Recognition}, 5333--5343.

\bibitem[{Shi et~al.(2020)Shi, Guo, Jiang, Wang, Shi, Wang, and Li}]{shi2020pv}
Shi, S.; Guo, C.; Jiang, L.; Wang, Z.; Shi, J.; Wang, X.; and Li, H. 2020.
\newblock Pv-rcnn: Point-voxel feature set abstraction for 3d object detection.
\newblock In \emph{Proceedings of the IEEE/CVF conference on computer vision
  and pattern recognition}, 10529--10538.

\bibitem[{Snavely, Seitz, and Szeliski(2006)}]{snavely2006photo}
Snavely, N.; Seitz, S.~M.; and Szeliski, R. 2006.
\newblock Photo tourism: exploring photo collections in 3D.
\newblock In \emph{ACM siggraph 2006 papers}, 835--846.

\bibitem[{Strang and Nguyen(1996)}]{strang1996wavelets}
Strang, G.; and Nguyen, T., eds. 1996.
\newblock \emph{Wavelets and filter banks}.
\newblock Wellesley: SIAM.

\bibitem[{Tancik et~al.(2022)Tancik, Casser, Yan, Pradhan, Mildenhall,
  Srinivasan, Barron, and Kretzschmar}]{tancik2022block}
Tancik, M.; Casser, V.; Yan, X.; Pradhan, S.; Mildenhall, B.; Srinivasan,
  P.~P.; Barron, J.~T.; and Kretzschmar, H. 2022.
\newblock Block-nerf: Scalable large scene neural view synthesis.
\newblock In \emph{Proceedings of the IEEE/CVF Conference on Computer Vision
  and Pattern Recognition}, 8248--8258.

\bibitem[{Wang et~al.(2024)Wang, Fang, Zhang, Xie, and
  Tian}]{wang2024gaussianeditor}
Wang, J.; Fang, J.; Zhang, X.; Xie, L.; and Tian, Q. 2024.
\newblock Gaussianeditor: Editing 3d gaussians delicately with text
  instructions.
\newblock In \emph{Proceedings of the IEEE/CVF Conference on Computer Vision
  and Pattern Recognition}, 20902--20911.

\bibitem[{Wang, Wang, and Qi(2024)}]{wang2024we}
Wang, Y.; Wang, J.; and Qi, Y. 2024.
\newblock WE-GS: An In-the-wild Efficient 3D Gaussian Representation for
  Unconstrained Photo Collections.
\newblock \emph{arXiv preprint arXiv:2406.02407}.

\bibitem[{Wang et~al.(2004)Wang, Bovik, Sheikh, and Simoncelli}]{wang2004image}
Wang, Z.; Bovik, A.~C.; Sheikh, H.~R.; and Simoncelli, E.~P. 2004.
\newblock Image quality assessment: from error visibility to structural
  similarity.
\newblock \emph{IEEE transactions on image processing}, 13(4): 600--612.

\bibitem[{Wen et~al.(2019)Wen, Zhang, Li, and Fu}]{wen2019pixel2mesh}
Wen, C.; Zhang, Y.; Li, Z.; and Fu, Y. 2019.
\newblock Pixel2mesh++: Multi-view 3d mesh generation via deformation.
\newblock In \emph{Proceedings of the IEEE/CVF international conference on
  computer vision}, 1042--1051.

\bibitem[{Williams(1983)}]{williams1983pyramidal}
Williams, L. 1983.
\newblock Pyramidal parametrics.
\newblock In \emph{Proceedings of the 10th annual conference on Computer
  graphics and interactive techniques}, 1--11.

\bibitem[{Wu et~al.(2015)Wu, Song, Khosla, Yu, Zhang, Tang, and
  Xiao}]{wu20153d}
Wu, Z.; Song, S.; Khosla, A.; Yu, F.; Zhang, L.; Tang, X.; and Xiao, J. 2015.
\newblock 3d shapenets: A deep representation for volumetric shapes.
\newblock In \emph{Proceedings of the IEEE conference on computer vision and
  pattern recognition}, 1912--1920.

\bibitem[{Xie et~al.(2024)Xie, Zong, Qiu, Li, Feng, Yang, and
  Jiang}]{xie2024physgaussian}
Xie, T.; Zong, Z.; Qiu, Y.; Li, X.; Feng, Y.; Yang, Y.; and Jiang, C. 2024.
\newblock Physgaussian: Physics-integrated 3d gaussians for generative
  dynamics.
\newblock In \emph{Proceedings of the IEEE/CVF Conference on Computer Vision
  and Pattern Recognition}, 4389--4398.

\bibitem[{Xu, Mei, and Patel(2024)}]{xu2024wild}
Xu, J.; Mei, Y.; and Patel, V.~M. 2024.
\newblock Wild-GS: Real-Time Novel View Synthesis from Unconstrained Photo
  Collections.
\newblock \emph{arXiv preprint arXiv:2406.10373}.

\bibitem[{Yan et~al.(2024)Yan, Lin, Zhou, Wang, Sun, Zhan, Lang, Zhou, and
  Peng}]{yan2024street}
Yan, Y.; Lin, H.; Zhou, C.; Wang, W.; Sun, H.; Zhan, K.; Lang, X.; Zhou, X.;
  and Peng, S. 2024.
\newblock Street gaussians for modeling dynamic urban scenes.
\newblock \emph{arXiv preprint arXiv:2401.01339}.

\bibitem[{Yang et~al.(2023)Yang, Zhang, Huang, Zhang, and Tan}]{yang2023cross}
Yang, Y.; Zhang, S.; Huang, Z.; Zhang, Y.; and Tan, M. 2023.
\newblock Cross-ray neural radiance fields for novel-view synthesis from
  unconstrained image collections.
\newblock In \emph{Proceedings of the IEEE/CVF International Conference on
  Computer Vision}, 15901--15911.

\bibitem[{Yu et~al.(2024)Yu, Chen, Huang, Sattler, and Geiger}]{yu2024mip}
Yu, Z.; Chen, A.; Huang, B.; Sattler, T.; and Geiger, A. 2024.
\newblock Mip-splatting: Alias-free 3d gaussian splatting.
\newblock In \emph{Proceedings of the IEEE/CVF Conference on Computer Vision
  and Pattern Recognition}, 19447--19456.

\bibitem[{Zhang et~al.(2024)Zhang, Wang, Wang, Li, Qin, and
  Wang}]{zhang2024gaussian-w}
Zhang, D.; Wang, C.; Wang, W.; Li, P.; Qin, M.; and Wang, H. 2024.
\newblock Gaussian in the Wild: 3D Gaussian Splatting for Unconstrained Image
  Collections.
\newblock \emph{arXiv preprint arXiv:2403.15704}.

\bibitem[{Zhang et~al.(2018)Zhang, Isola, Efros, Shechtman, and
  Wang}]{zhang2018unreasonable}
Zhang, R.; Isola, P.; Efros, A.~A.; Shechtman, E.; and Wang, O. 2018.
\newblock The unreasonable effectiveness of deep features as a perceptual
  metric.
\newblock In \emph{Proceedings of the IEEE conference on computer vision and
  pattern recognition}, 586--595.

\bibitem[{Zhou et~al.(2024{\natexlab{a}})Zhou, Shao, Xu, Bai, Qiu, Liu, Wang,
  Geiger, and Liao}]{zhou2024hugs}
Zhou, H.; Shao, J.; Xu, L.; Bai, D.; Qiu, W.; Liu, B.; Wang, Y.; Geiger, A.;
  and Liao, Y. 2024{\natexlab{a}}.
\newblock Hugs: Holistic urban 3d scene understanding via gaussian splatting.
\newblock In \emph{Proceedings of the IEEE/CVF Conference on Computer Vision
  and Pattern Recognition}, 21336--21345.

\bibitem[{Zhou et~al.(2024{\natexlab{b}})Zhou, Chang, Jiang, Fan, Zhu, Xu,
  Chari, You, Wang, and Kadambi}]{zhou2024feature}
Zhou, S.; Chang, H.; Jiang, S.; Fan, Z.; Zhu, Z.; Xu, D.; Chari, P.; You, S.;
  Wang, Z.; and Kadambi, A. 2024{\natexlab{b}}.
\newblock Feature 3dgs: Supercharging 3d gaussian splatting to enable distilled
  feature fields.
\newblock In \emph{Proceedings of the IEEE/CVF Conference on Computer Vision
  and Pattern Recognition}, 21676--21685.

\bibitem[{Zhou et~al.(2024{\natexlab{c}})Zhou, Lin, Shan, Wang, Sun, and
  Yang}]{zhou2024drivinggaussian}
Zhou, X.; Lin, Z.; Shan, X.; Wang, Y.; Sun, D.; and Yang, M.-H.
  2024{\natexlab{c}}.
\newblock Drivinggaussian: Composite gaussian splatting for surrounding dynamic
  autonomous driving scenes.
\newblock In \emph{Proceedings of the IEEE/CVF Conference on Computer Vision
  and Pattern Recognition}, 21634--21643.

\bibitem[{Zwicker et~al.(2001)Zwicker, Pfister, Van~Baar, and
  Gross}]{zwicker2001ewa}
Zwicker, M.; Pfister, H.; Van~Baar, J.; and Gross, M. 2001.
\newblock EWA volume splatting.
\newblock In \emph{Proceedings Visualization, 2001. VIS'01.}, 29--538. IEEE.

\end{thebibliography}

\clearpage
\appendix

\twocolumn[
\begin{@twocolumnfalse}
\section*{\centering{\LARGE Supplementary Material for \\ \emph{Micro-macro Wavelet-based Gaussian Splatting for 3D Reconstruction from Unconstrained Images\\[25pt]}}}
\end{@twocolumnfalse}
]

\begin{table*}[t!]
    \centering
    \begin{tabular}{cccc}
    \toprule
        Method & \textit{Brandenburg Gate} & \textit{Sacre Coeur} & \textit{Trevi Fountain}\\
    \midrule
        NeRF-W & 0.0518 & 0.0514 & 0.0485 \\ 
        Ha-NeRF & 0.0489 & 0.0497 & 0.0498\\
        CR-NeRF & 0.0445 & 0.0447 & 0.0446\\
        WildGaussians & 36.10 & 40.42 & 18.54\\
        GS-W & \underline{54.49} & \underline{60.31} & \underline{39.99}\\
        ours & \textbf{64.43} & \textbf{92.08} & \textbf{61.01}\\
    \bottomrule
    \end{tabular}
    \caption{We compared the rendering speed on three datasets with a resolution of $800 \times 800$ using a single RTX 3090 GPU, measuring performance in FPS (frames per second). \textbf{Bold} and \underline{underlined} values correspond to the best and the second-best value, respectively.}
    \label{tab:speed}
\end{table*}

\begin{table*}[t!]
\centering
\begin{tabular}{cccccccccc}
\toprule
\multirow{2}{*}{Method} & \multicolumn{3}{c}{\textit{Brandenburg Gate}} & \multicolumn{3}{c}{\textit{Sacre Coeur}} & \multicolumn{3}{c}{\textit{Trevi Fountain}} \\ 
\cmidrule(lr){2-4} \cmidrule(lr){5-7} \cmidrule(lr){8-10} 
& PSNR $\uparrow$ & SSIM $\uparrow$ & LPIPS $\downarrow$ & PSNR $\uparrow$ & SSIM $\uparrow$ & LPIPS $\downarrow$ & PSNR $\uparrow$ & SSIM $\uparrow$ & LPIPS $\downarrow$ \\
\midrule
$M=0$ & 28.88  & 0.939  & 0.055 & 24.35  & 0.895 & 0.075 & 23.38  & 0.810 & 0.127 \\
$M=1$ & \textbf{29.37} & \textbf{0.942} & \textbf{0.052} & \textbf{24.64} & \textbf{0.897} & \textbf{0.073} & \textbf{24.07} & \textbf{0.821} & \textbf{0.120} \\
$M=2$ & 28.94  & 0.938  & 0.054 & 24.41  & 0.897  & 0.076 & 23.40 &  0.810   & 0.127 \\
\bottomrule
\end{tabular}
\caption{Quantitative results across three datasets for different values of \(M\), where \(M\) represents the number of downsampling steps for the lowest dimension in the multi-scale resolution, related to the dimension of the wavelet transform. We selected \(M=1\) to balance efficiency and effectiveness. \textbf{Bold} values correspond to the best value.}
\label{tab:feat_sampling_all}
\end{table*}

\begin{table*}[t!]
\centering
\begin{tabular}{cccccccccc}
\toprule
\multirow{2}{*}{Method} & \multicolumn{3}{c}{\textit{Brandenburg Gate}} & \multicolumn{3}{c}{\textit{Sacre Coeur}} & \multicolumn{3}{c}{\textit{Trevi Fountain}} \\ 
\cmidrule(lr){2-4} \cmidrule(lr){5-7} \cmidrule(lr){8-10} 
& PSNR $\uparrow$ & SSIM $\uparrow$ & LPIPS $\downarrow$ & PSNR $\uparrow$ & SSIM $\uparrow$ & LPIPS $\downarrow$ & PSNR $\uparrow$ & SSIM $\uparrow$ & LPIPS $\downarrow$ \\
\midrule
$k_s=1$ & \textbf{29.37} & \textbf{0.942} & \textbf{0.052} & \textbf{24.64} & \textbf{0.897} & \textbf{0.073} & \textbf{24.07}  & \textbf{0.821}  & \textbf{0.120} \\
$k_s=2$ & 29.02  & 0.938  & 0.057 & 23.96  & 0.885 &  0.081 & 23.74 &  0.815 &  0.124\\
$k_s=3$ & 29.01  &   0.941  & 0.053 & 24.40 &  0.893  & 0.075 &  23.81 & 0.811  & 0.126\\
\bottomrule
\end{tabular}
\caption{Quantitative results on three datasets for different $k_s$ values, where \( k_s \) represents the number of samples within the cross-section of each conical frustum. We Choose $k_s=1$ to achieve efficiency and effectiveness . \textbf{Bold} values correspond to the best value.}
\label{tab:ks}
\end{table*}

\section{More Implementation Details}
Below, we provide a detailed overview of all hyper-parameter settings in MW-GS and the network parameters, as well as the computational infrastructure used for the experiments. 

We develop our method based on the original implementation of Scaffold-GS. In our experiments, we set the number of Gaussians per voxel to \( k =10\)  , the number of samples within the frustum to \( k_s = 1 \), the maximum wavelet dimension to \( m = 1 \), the intrinsic feature dimension to \( n_v = 48 \), the refined appearance feature dimension to \( n_r = 32 \), and the global appearance feature dimension to \( n_g = 16 \). Furthermore, the learning rates of \(nc_i\) and \(bc_i\) decrease from \(1\times 10^{-4}\) to \(1\times 10^{-5}\). We trained our networks using the Adam optimizer, with \(\lambda_{\text{ssim}} = 0.2\), \(\lambda_{1} = 0.8\), and \(\lambda_{\text{vm}} = 0.15\). Other hyper-parameters are set according to the guidelines of Scaffold-GS.

We constructed the encoder of the UNet to extract image features by utilizing all the modules from the pre-trained ResNet-18 up to the penultimate layer, just before the AdaptiveAvgPool2d layer. The MLP used to generate the global appearance feature \(MLP^G\), has a single hidden layer with the number of hidden units set to \( 2n_g \), and it ultimately outputs the global appearance feature. The UNet decoder used to generate feature maps for detailed feature modeling consists of four upsampling convolutional modules, equipped with residual connections that follow the original UNet design. The dimensions are then adjusted to \( n_r \) through a final convolutional layer. Meanwhile, the UNet decoder for generating the visibility map includes three upsampling convolutional modules and does not include residual connections. The learning rate for the networks mentioned above, used for appearance feature extraction, is gradually decreased from \(1\times 10^{-4}\) to \(1\times 10^{-6}\). Given that Gaussian training is performed on a single image at a time, our decoder is equipped only with ReLU activation functions, without batch normalization. Additionally, the batch normalization layers in the pretrained encoder are frozen.

The proposed Hierarchical Residual Fusion Network (HRFN) consists of four MLPs, denoted as \( \mathcal{M}^H = \{ \mathcal{M}^H_1, \mathcal{M}^H_2, \mathcal{M}^H_3, \mathcal{M}^H_4 \} \), with the number of hidden units for each set as follows: \{128, 96\}, \{96, 64\}, \{48, 48\}, and \{48\}. The learning rate for the HRFN gradually decreases from \(5\times 10^{-4}\) to \(5\times 10^{-5}\).

Our framework and experiments were implemented using Python version 3.9.18 and the PyTorch library version 1.12.1, running on Ubuntu 22.04. Training was conducted on a single Nvidia RTX 3090 GPU for 60,000 steps.

\section{Render Speed}
To compare the rendering speeds of different methods during inference, we conducted experiments on three datasets with an image resolution of $800 \times 800$, using a single RTX 3090 GPU to calculate the average rendering time per image. The overall inference time includes the time required to extract features from the reference image for Ha-NeRF, CR-NeRF, GS-W, and our method. As shown in the Tab. \ref{tab:speed}, our method not only ensures fine-grained modeling of image appearance but also achieves excellent rendering speed, being nearly 1.5 times faster than existing Gaussian-based methods.

\section{Extended Component Analysis}
\subsection{Analysis of Wavelet Dimension}
In our method, \( M \) represents the max dimension of the discrete wavelet decomposition, which corresponds to the number of times the feature map is downsampled. We conducted experiments on three datasets, setting \( M \) to 0, 1, and 2. The experimental results are shown in Tab. \ref{tab:feat_sampling_all}. We observed that when \( M = 1 \), sampling at both 1x resolution and 0.5x resolution, the performance on the test set is optimal. Increasing the number of downsampling steps, such as performing downsampling at 1x resolution, 0.5x resolution, and 0.25x resolution, does not bring additional performance improvements and incurs higher computational costs. Therefore, we reasonably set \( M \) to 1.

\subsection{Analysis of Sampling Number}
In our method, \( k_s \) represents the number of samples within the cross-section of each conical frustum. We conducted experiments across three different scenes to determine the optimal value for \( k_s \), testing settings of \( k_s = 1 \), \( k_s = 2 \), and \( k_s = 3 \). The results are shown in Tab. \ref{tab:ks}. Surprisingly, we observed that when \( k_s = 1 \), the performance on the test set was remarkably high. Increasing \( k_s \) did not lead to further performance improvements and instead resulted in higher computational costs. We think that additional sampling might cause the features of 3D points along the same ray to converge towards a common mean, reducing diversity and ultimately degrading performance. Therefore, we reasonably set \( k_s \) to 1.
\section{Limitation}
Similar to previous methods, MW-GS still struggles to recover fine ground textures (such as roads or sidewalks), resulting in blurred regions in these areas. We believe this is largely due to the lack of sufficient supervision in these parts, making it meaningful to use advanced diffusion models to fill in the missing information. Additionally, since the transient masks are learned in an unsupervised manner, they sometimes obscure areas with unusual appearances (which are difficult to model but easy to mask). While using more advanced segmentation networks could yield more accurate transient masks, benefiting appearance modeling and geometric reconstruction, the speed of these segmentation networks would significantly limit the training of 3DGS.

Therefore, further development of more advanced and efficient transient masking techniques and the use of advanced diffusion models to complete ground textures are promising and meaningful directions for future work.

\end{document}